%% file: main.tex
\documentclass[10pt,twocolumn,letterpaper]{article}
\usepackage[accsupp]{axessibility}
\usepackage{cvpr}              % To produce the CAMERA-READY version
\input{preamble}

\definecolor{cvprblue}{rgb}{0.21,0.49,0.74}
\usepackage[pagebackref,breaklinks,colorlinks,citecolor=cvprblue]{hyperref}

\usepackage{xspace}
\usepackage{multirow}
\usepackage{xcolor}
\usepackage{times}
\usepackage{epsfig}
\usepackage{amsmath}
\usepackage{amssymb}
\usepackage{booktabs}
\usepackage{bbding}
\usepackage{enumitem}
\usepackage{tablefootnote}
\usepackage{makecell}
\usepackage{nicematrix}

\usepackage{array,etoolbox}
\preto\tabular{\setcounter{magicrownumbers}{0}}
\newcounter{magicrownumbers}
\def\rownumber{}

\usepackage{color, colortbl}
\definecolor{Gray}{gray}{0.85}

\makeatletter
\newcommand\figcaption{\def\@captype{figure}\caption}
\newcommand\tabcaption{\def\@captype{table}\caption}
\makeatother
\definecolor{bluegray}{rgb}{0.4, 0.6, 0.8}
\definecolor{atomictangerine}{rgb}{1.0, 0.6, 0.4}
\definecolor{bananayellow}{rgb}{1.0, 0.88, 0.21}
\definecolor{amethyst}{rgb}{0.6, 0.4, 0.8}
\definecolor{newblue}{rgb}{0.45, 0.56, 0.80}

\newcommand\blfootnote[1]{
  \begingroup
  \renewcommand\thefootnote{}\footnote{#1}
  \addtocounter{footnote}{-1}
  \endgroup
}

\newcommand{\system}{OmniViD\xspace}
\def\confName{CVPR}
\def\confYear{2024}

\title{\system: A Generative Framework for Universal Video Understanding}

\author{Junke Wang$^{1,2}$,~Dongdong Chen$^{3}$,~Chong Luo$^{4}$,~Bo He$^{5}$,~Lu Yuan$^{3}$,~Zuxuan Wu$^{1,2\dagger}$,~Yu-Gang Jiang$^{1,2}$
\\[0.5em]
$^{1}$Shanghai Key Lab of Intell. Info. Processing, School of CS, Fudan University \\
$^{2}$Shanghai Collaborative Innovation Center of Intelligent Visual Computing \\
$^{3}$Microsoft Cloud + AI, $^{4}$Microsoft Research Asia, $^{5}$University of Maryland, College Park  
}

\begin{document}
\maketitle

\input{sec/0_abstract}

\input{sec/1_intro}
\input{sec/2_related}
\input{sec/3_method}

\input{sec/4_experiment}
\input{sec/5_conclusion}

%\clearpage

{
    \small
    \bibliographystyle{ieeenat_fullname}
    \bibliography{main}
}

%\appendix

\end{document}

%% file: preamble.tex
\usepackage[dvipsnames]{xcolor}

%% file: sec/0_abstract.tex
\begin{abstract}
The core of video understanding tasks, such as recognition, captioning, and tracking, is to automatically detect objects or actions in a video and analyze their temporal evolution. Despite sharing a common goal, different tasks often rely on distinct model architectures and annotation formats. In contrast, natural language processing benefits from a unified output space, \textit{i.e.}, text sequences, which simplifies the training of powerful foundational language models, such as GPT-3, with extensive training corpora. Inspired by this, we seek to unify the output space of video understanding tasks by using languages as labels and additionally introducing \textit{time} and \textit{box} tokens. In this way, a variety of video tasks could be formulated as video-grounded token generation. This enables us to address various types of video tasks, including classification (such as action recognition), captioning (covering clip captioning, video question answering, and dense video captioning), and localization tasks (such as visual object tracking) within a fully shared encoder-decoder architecture, following a generative framework. Through comprehensive experiments, we demonstrate such a simple and straightforward idea is quite effective and can achieve state-of-the-art or competitive results on seven video benchmarks, providing a novel perspective for more universal video understanding. Code is available at \href{https://github.com/wangjk666/OmniVid}{https://github.com/wangjk666/OmniVid}.
\blfootnote{$^{\dagger}$Corresponding author.}
\end{abstract}

%% file: sec/1_intro.tex
\section{Introduction}
\label{sec:intro}
In recent years, the proliferation of video content across various applications, such as online education and live streaming, has profoundly impacted our daily lives. Videos have evolved into a captivating and immersive medium for information delivery, emphasizing the pressing demand for the development of automated algorithms capable of understanding the actions~\cite{kay2017kinetics}, events~\cite{krishna2017dense}, and moving objects~\cite{smeulders2013visual} within video sequences. As a result, the field of video understanding has undergone significant expansion and encompassed a diverse range of tasks, including action recognition~\cite{simonyan2014two,wu2015modeling,feichtenhofer2019slowfast,bertasius2021space,li2022mvitv2,liu2022video}, video captioning~\cite{gao2017video,chen2019deep,lin2022swinbert}, and object tracking~\cite{yilmaz2006object,bertinetto2016fully,chen2021transformer,wang2023look}. 

For a long period, research in video understanding has adopted a task-specific paradigm, \ie, designing specialized architectures and loss functions to cater to the unique requirements of different tasks and benchmarks~\cite{kay2017kinetics,xu2016msr,caba2015activitynet,muller2018trackingnet,he2022asm,he2023align}. Despite the promising results with high-capacity deep neural networks, these methods~\cite{fan2021multiscale,tian2023resformer,yang2023vid2seq,wu2024building} are tailored for a particular objective and less adaptable to deployment in scenarios of diverse needs. To mitigate this issue, video foundation models~\cite{wang2022omnivl,wang2022internvideo,xing2023vidiff,wang2023chatvideo,he2024malmm}, have gained emerging attention for their impressive performance across a broad spectrum of video tasks and potential in realizing the vision of Artificial General Intelligence (AGI). However, while generic spatial-temporal representations can be learned with these models, adapting them to different downstream tasks often requires carefully designing and fine-tuning task-specific heads.

\begin{figure*}[t]
\centering
\includegraphics[width=\linewidth]{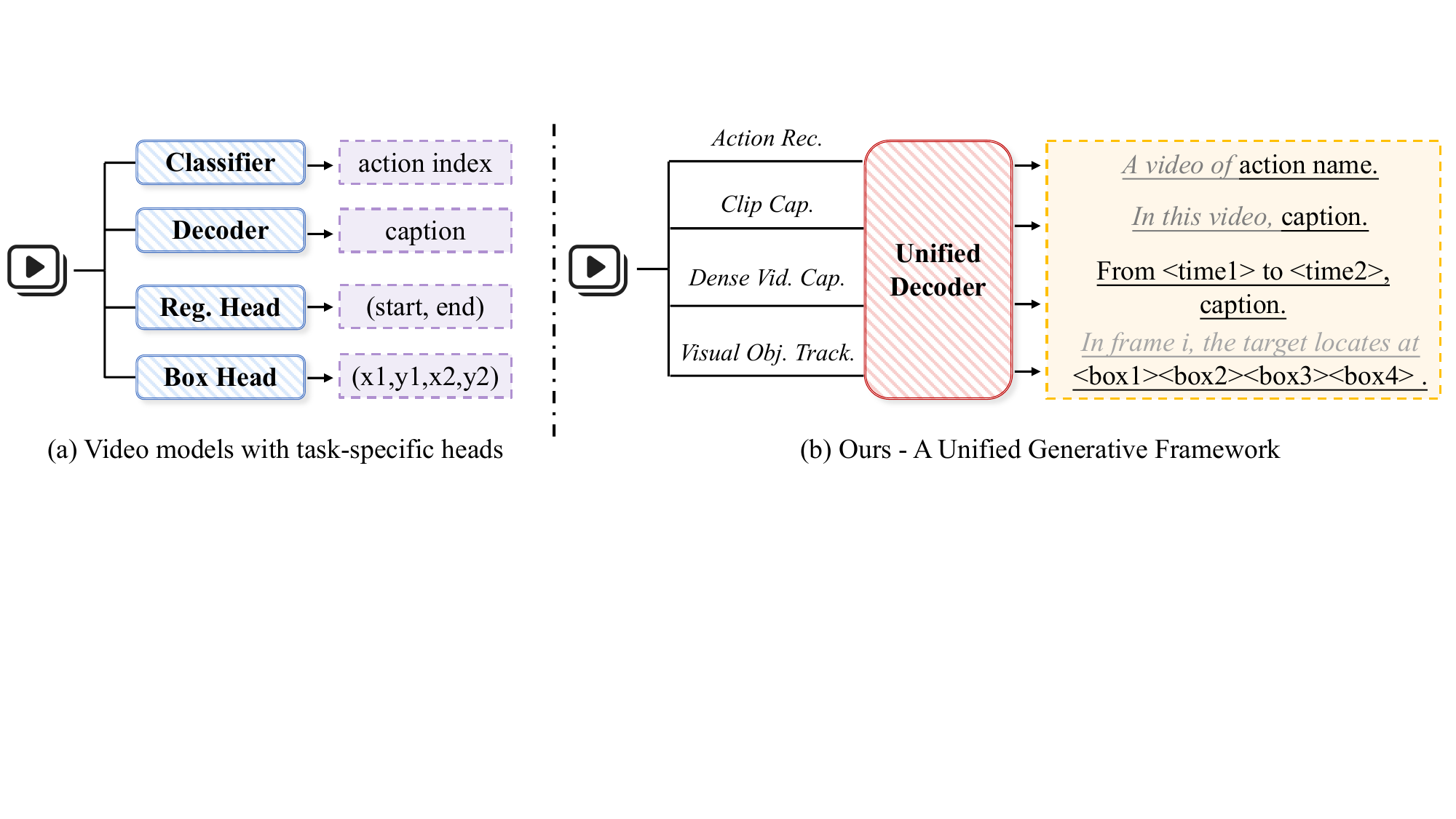}
\vspace{-0.2in}
\caption{A conceptual comparison between existing video models and \system.}
\vspace{-0.15in}
\label{fig:sequence}
\end{figure*}
        
In this paper, we posit such limitation originates from the diversified annotations for different video tasks, \eg, a set of action categories for action recognition~\cite{simonyan2014two,wu2015modeling,carreira2017quo}, sentences for captioning~\cite{gao2017video,lin2022swinbert}, and continuous segments (coordinates) for events (object) localization~\cite{mun2020local,chen2021end,chen2021transformer}. This naturally necessitates task-specific designs for better optimization. In contrast, different tasks in natural language processing (NLP) enjoy a sharable output space, \ie, text sequences, which promotes the development of large language models, such as GPT~\cite{radford2018improving,radford2019language} and Llama~\cite{touvron2023llama,touvron2023llama2,jiao2024lumen}. Drawing inspiration from this, we 
leverage word tokens in natural languages to represent semantic information that is important for coarse-grained tasks like action recognition, video captioning, and video question answering, and additionally introduce special \textit{time tokens} and \textit{box tokens} that provide localization capabilities in both spatial and temporal dimensions, particularly useful for fine-grained tasks like dense video captioning and visual object tracking. With such an enriched vocabulary that consists of word, time, and box tokens, the output format, as well as training objectives of different tasks, can be well unified. Please refer to Figure~\ref{fig:sequence} for a better illustration.

With this in mind, we present \system, a generative framework that approaches various video tasks as a language modeling task conditioned on video inputs. \system adopts an encoder-decoder architecture, where a dedicated video encoder and a language encoder are employed to extract the multimodal features from diverse inputs. Considering the remarkable redundancy in video data, we propose a lightweight MQ-former to enhance the efficiency of video representations for subsequent modeling. The MQ-former utilizes three types of learnable queries, \ie, content, sentence, and box queries, to aggregate the frame features from the video encoder through cross-attention. Finally, a token decoder is applied to generate a token sequence from the above vocabulary.

We validate the effectiveness of \system on five representative video tasks, including action recognition, clip captioning, video question answering, dense video captioning, and visual object tracking. The results demonstrate that \system achieves new state-of-the-art or at least competitive results on the prevalent video benchmarks. For example, using VideoSwin-Base~\cite{liu2022video} as the video encoder, we achieve state-of-the-art performance on action recognition (83.6\% top1 accuracy on Kinetics-400~\cite{kay2017kinetics}), clip captioning (56.6 on MSRVTT~\cite{xu2016msr} in terms of CIDEr ), video question answering (42.3\% accuracy on MSRVTT~\cite{xu2016msr}), dense video captioning (5.6 on ActivityNet~\cite{caba2015activitynet} in terms of SODA\_c), and visual object tracking (88.9 on TrackingNet~\cite{muller2018trackingnet} in terms of normalized precision). For the first time, video tasks of different modalities and granularity can be supported by a single framework.

%% file: sec/2_related.tex
\section{Related Work}
\label{sec:related}
\begin{figure*}[t]
\begin{minipage}[b]{.57\textwidth}
\centering
\includegraphics[scale=0.41]{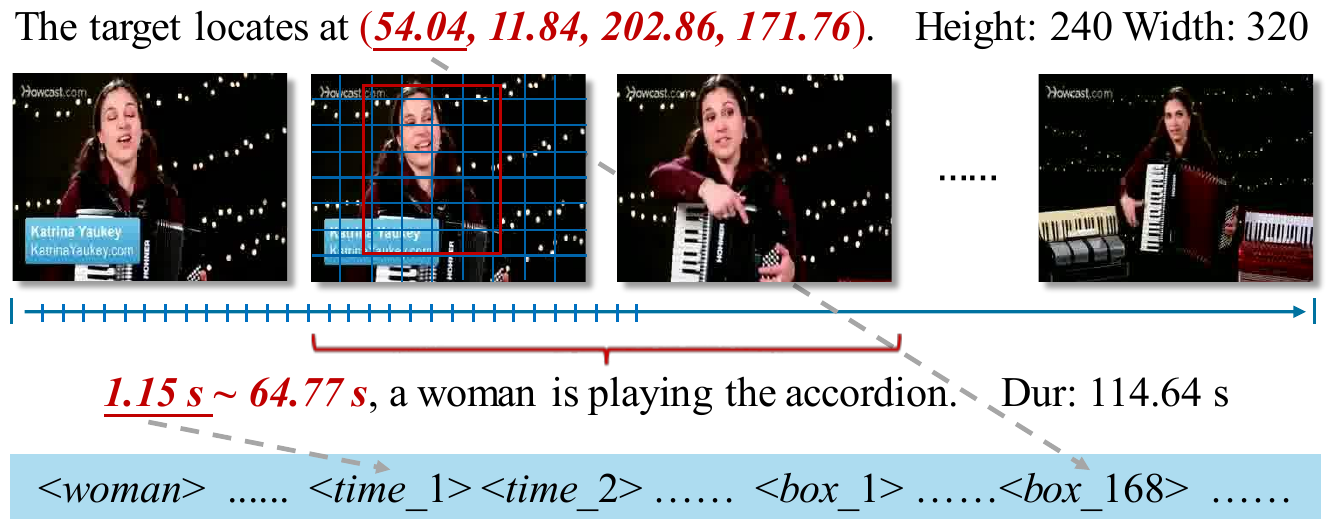}
\figcaption{Illustration of the \textit{time tokens} and \textit{box tokens} in \system.} 
\label{fig:vocabulary}
\end{minipage}
~
\begin{minipage}[b]{0.43\textwidth}
\centering
\tabcaption{Input $\&$ output of different video tasks. S/B/W/T: Sentence / Box / Word / Time, Pro. / Tok.: Prompt / Token.}
\vspace{-0.1in}
\resizebox{\linewidth}{!}{
\begin{tabular}{lc | cc | ccc}
\toprule
\multirow{2}*{\textbf{Task}} && \multicolumn{2}{c|}{\textbf{Input}} & \multicolumn{3}{c}{\textbf{Target}} \\
~ && S Pro. & B Pro. & W Tok. & T Tok. & B Tok. \\
\midrule
AR && \XSolidBrush & \XSolidBrush & \Checkmark & \XSolidBrush & \XSolidBrush \\
\midrule
CC && \XSolidBrush & \XSolidBrush & \Checkmark & \XSolidBrush & \XSolidBrush \\
ViQA && \Checkmark & \XSolidBrush & \Checkmark & \XSolidBrush & \XSolidBrush \\
DVP && \XSolidBrush & \XSolidBrush & \Checkmark & \Checkmark & \XSolidBrush \\
\midrule
VOT && \XSolidBrush & \Checkmark & \XSolidBrush & \XSolidBrush & \Checkmark \\
\bottomrule
\end{tabular}}
\label{tab:io}
\end{minipage}
\vspace{-0.25in}
\end{figure*}

\subsection{Task-specific Methods for Video Understanding}
\vspace{-0.05in}
Task-specific video understanding models could be roughly divided into classification, captioning, and localization approaches. Video action recognition is the most representative classification task in the video domain, which aims to recognize human actions in a video. Existing methods, including both CNN-based~\cite{feichtenhofer2016convolutional,kay2017kinetics,feichtenhofer2019slowfast,he2020gta,ma2022rethinking} and Transformer-based models~\cite{bertasius2021space,fan2021multiscale,liu2022video}, widely encode the action labels as one-hot vectors and employ cross-entropy loss for supervised training. Captioning tasks, on the other hand, typically generate a textual description for a video clip~\cite{zhou2018end,zhou2018towards,lin2022swinbert} or an untrimmed long video~\cite{iashin2020multi,wang2021end,yang2023vid2seq} with a text decoder like BERT~\cite{kenton2019bert}. It is worth noting that captioning long videos involves the additional challenge of temporal event localization within the video, making it a more complex task. We categorize the open-ended video question answering~\cite{lei2018tvqa,le2020hierarchical,li2022invariant} as a specific type of captioning task due to the consistent output format between them. Localization tasks, represented by visual object tracking~\cite{wang2015visual,chen2021transformer,cui2022mixformer}, estimate the trajectory of a target object in a video sequence given its position in the first frame. Following the practice in object detection~\cite{he2017mask,girshick2015fast,carion2020end}, a box head is oftentimes adopted to regress the coordinates of the tracking object. In summary, divergent prediction heads have been developed in various video tasks to adapt to the specific format of annotations, which poses a challenge to derive a unified solution. In this paper, we rethink the design of a universal video understanding framework from a novel perspective, \ie, redefining an output space that could be shared by different video tasks. Within this unified space, the development of general architectures and training objectives become distinctly feasible.

\vspace{-0.05in}
\subsection{Unified Video Models}
\vspace{-0.05in}
Recently, researchers have undertaken prominent efforts to unify video tasks within specific domains. OmniVL~\cite{wang2022omnivl} and InterVideo~\cite{wang2022internvideo} represent significant strides in the realm of video-language pretraining, which are pre-trained on large-scale video-text data and achieve superior results on multimodal video tasks like text-to-video retrieval and video captioning. Beyond these advancements, UNLoc~\cite{yan2023unloc} and UniVTG~\cite{qinghong2023univtg} have sought to tackle a diverse array of temporal localization tasks within a single framework. They accomplish this by simultaneously predicting saliency scores and boundary offsets for each frame (clip). Compared to video-language and temporal localization, spatial localization in the video domain, \ie, tracking, is more fragmented in terms of task definition, model architecture, and benchmarks. Unicorn~\cite{yan2022towards} marks a significant step forward by employing a fully shared CNN-based encoder and box head for various tracking tasks, utilizing a target before distinguishing between them. Subsequently, with the prominent success of vision transformer~\cite{carion2020end}, OmniTracker~\cite{wang2023omnitracker} and UNINEXT~\cite{yan2023universal} push the boundaries of unification in tracking models by incorporating Transformer-based detectors. Despite the achievements of these approaches, they are still constrained by task-specific heads, leaving considerable space for greater unification of video understanding. To address this limitation, we unify diverse tasks with a sharable output space and address them with a fully shared generative framework.

\begin{figure*}[t]
  \centering
  \includegraphics[width=\linewidth]{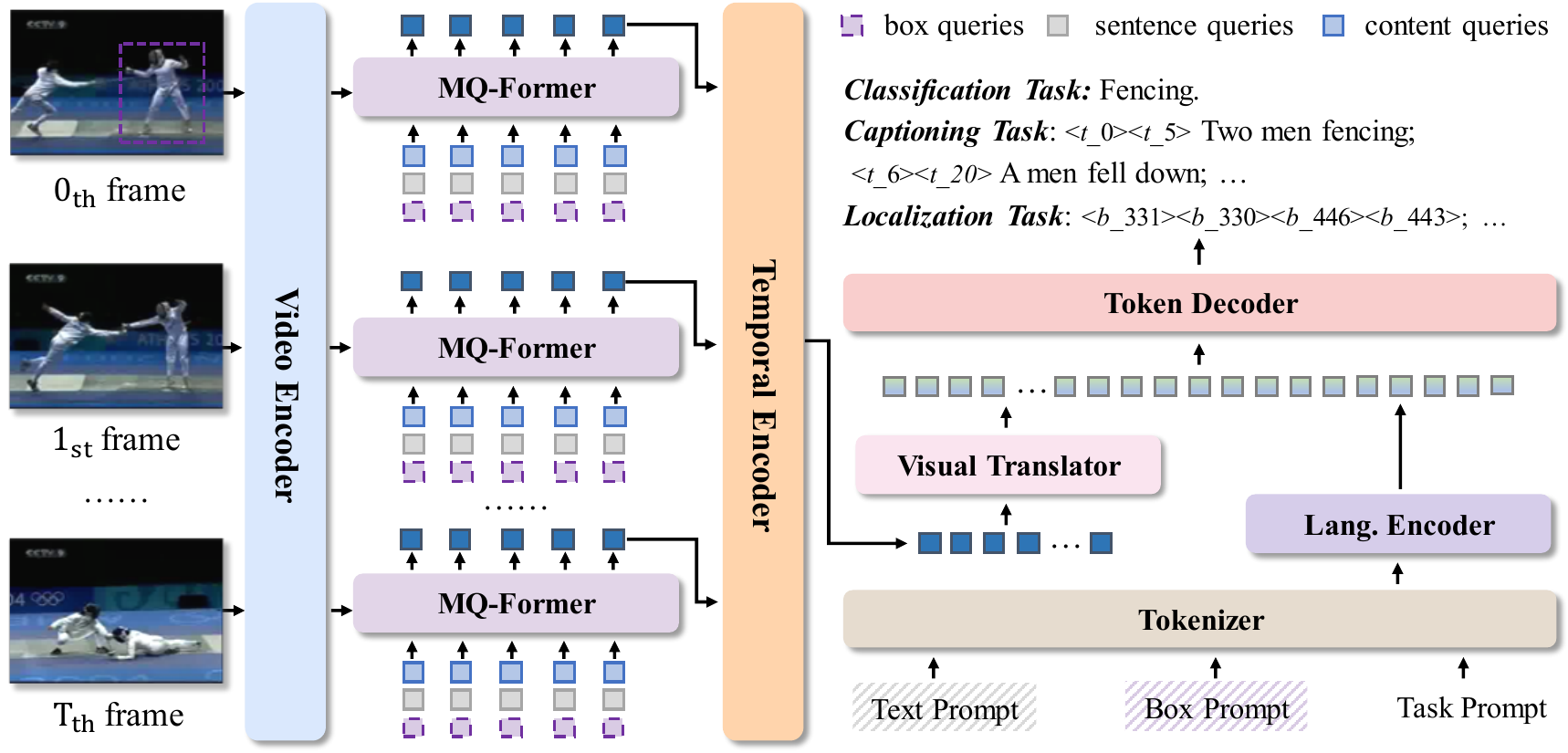}
  \vspace{-0.1in}
  \caption{Architecture of \system. The Mixed Q-former aggregates the frame features into three types of queries, \ie, \textcolor{newblue}{content queries}, \textcolor{gray}{text queries}, and \textcolor{amethyst}{box queries}. After that, the queries obtained from different frames are input to a temporal encoder for temporal modeling. Finally, the token decoder generates a sequence of tokens conditioned on the multimodal inputs. }
  \label{fig:architecture}
  \vspace{-0.1in}
\end{figure*}

\vspace{-0.05in}
\subsection{Autoregress Modeling in Computer Vision}
\vspace{-0.05in}
AutoRegressive modeling~\cite{yang2019xlnet} is a statistical modeling technique that predicts the current state of a sequence based on historical observations, which has achieved remarkable success in the field of natural language processing (NLP)~\cite{chowdhary2020natural} and time series analyasis~\cite{meek2002autoregressive,feigelson2018autoregressive}. Inspired by this, researchers in the vision community have also attempted to explore its potential for visual understanding. Pix2SeqV1$\&$V2~\cite{chen2021pix2seq,chen2022unified} expand the textual vocabulary with quantized image coordinates. With this, they address several fundamental image tasks, \eg, object detection, and image captioning, in a unified autoregressive manner. Following this idea, ARTrack~\cite{wei2023autoregressive} and SeqTrack~\cite{chen2023seqtrack} further support the visual object tracking task. VisionLLM~\cite{wang2023visionllm}, on the other hand, directly builds vision-centric frameworks upon pre-trained LLMs, with the hope of transferring their knowledge to visual understanding with minimal resource overhead. In this work, we leverage autoregressive modeling to the design of a universal video understanding framework. In addition to the expansion to temporal localization tasks with unique \textit{time tokens}, our method also explores the advantages of autoregressive modeling for a universal video understanding framework for the first time.

%% file: sec/3_method.tex
\section{Method}
\label{sec:method}
Our primary objective is to design a universal framework that accommodates a diverse set of video understanding tasks. To accomplish this, we expand upon the vocabulary commonly used in language models~\cite{brown2020language,lewis2019bart} by introducing unique \textit{time tokens} and \textit{box tokens}. This augmentation allows us to represent the output of various video tasks as a token sequence within a shared vocabulary. Building upon this foundation, we further present \system, a generative framework that conceptualizes video tasks as a process for generating tokens grounded in the video content.

Given a video $\mathcal{V}$ that lasts tens of seconds to multiple minutes, we sample a sequence of frames $[\textit{X}_{1}, \textit{X}_{2}, ..., \textit{X}_{T}]$ from it. For video question answering, a question regarding the visual content is given, while for visual object tracking, the bounding box of the target object in the first frame is specified by the user. Below we first introduce how to perform tokenization for different video tasks with the above vocabulary in Sec.~\ref{subsec:tokenization}, and then present the architecture of \system in Sec.~\ref{subsec:arch}. Finally, we elaborate on the unified training and inference pipeline in Sec.~\ref{subsec:obj}.

\subsection{Unified Vocabulary for Video Understanding}
\label{subsec:tokenization}
In video understanding, various tasks necessitate diverse inputs and outputs according to specific settings and requirements. To establish a cohesive output space that could be shared by different video tasks, we supplement the word tokens in language vocabulary with special \textit{time tokens} and \textit{box tokens}, by discretizing the timestamps and the coordinates along the temporal and spatial dimensions, respectively (see Figure~\ref{fig:vocabulary}). 

With the enriched vocabulary, the input and target sequences for the training of \system can be generated in the following manner:
\begin{itemize}
    \vspace{0.01in}
    \item \textbf{Action Recognition}: the input only includes a task prompt $\textit{p}_{task}$, \ie, ``action recognition'', and the target is the ground-truth action name, \eg, ``dancing ballet''.
    \vspace{0.01in}
    \item \textbf{Clip Captioning}: similar to action recognition, the only difference lies in the target sequence becomes a longer description, \eg, ``a clip showing a computer screen''.
    \item \textbf{Video Question-Answering}: the input includes both the task prompt and the question $\textit{p}_{sen}$, \eg, ``What is the video doing?'', while the target is the answer to that question, \eg, ``fencing competition''. 
    \item \textbf{Dense Video Captioning}: the expected output is a set of events $\{e_{i}\}_{i=1}^{E}$ happening in the given video. In order to facilitate the model to learn the correspondence between timestamps and visual contents, we define a triplet for the $i$-th event $e_{i}$: $e_{i} = \left \langle t^{start}_{i}, t^{dur}_{i}, s \right \rangle$, where $t^{start}_{i}$ and $t^{dur}_{i}$ denote the start and duration time token, and $s$ represents the description for the event~\cite{yang2023vid2seq}. The target sequence is constructed by concatenating the triplets of all the events.
    \item \textbf{Visual Object Tracking}: we take the task prompt and the discrete representation of the bounding box in the first frame, $\textit{p}_{box}$, as input, and employ the \textit{box tokens} in the following frames as target. Given a bounding box (x1, y1, x2, y2) on an $H \times W$ image, the tokenized representation is ($\left \langle box\_\lfloor \mathrm{x1}/W \rfloor \right \rangle$, $\left \langle box\_\lfloor \mathrm{y1}/H \rfloor \right \rangle$, $\left \langle box\_ \lfloor \mathrm{x2}/W \rfloor \right \rangle$, $\left \langle box\_ \lfloor \mathrm{y2}/H \rfloor \right \rangle$).
\end{itemize}
\vspace{0.02in}
The input and target sequence for different video tasks are summarized in Table~\ref{tab:io}. 

\subsection{Unified Architecture}
\label{subsec:arch}
\system follows an encoder-decoder architecture, which first extracts the video features $F \in \mathcal{R}^{T^{f} \times H^{f} \times W^{f} \times C^{f}}$ from $\{\textit{X}_{t}\}_{t=1}^{T}$ with a video encoder, where $T^{f}$ and $H^{f}\times W^{f}$ denote the temporal and spatial resolution and $C^{f}$ is the feature dimension. For visual object tracking, we replace the first frame with the cropped template, following the common practice~\cite{wang2015visual,bertinetto2016fully,chen2021transformer}. A language encoder is also adopted to transform three types of prompts, $\textit{p}_{task}$, $\textit{p}_{sen}$, $\textit{p}_{box}$ to the prompt embeddings $G_{task}$, $G_{sen}$, $G_{box}$, and then concatenate them as the textual feature $G \in \mathcal{R}^{L^{g} \times C^{g}}$ along the sequence dimension. Based on the multimodal inputs, \system produces a sequence of tokens in the above vocabulary. The overall framework is illustrated in Figure~\ref{fig:architecture}. 

\vspace{0.02in}
\noindent \textbf{MQ-former.} In order to encode the video features into a more efficient representation, we further propose a MQ-former to aggregate them into a set of learnable queries. Ours MQ-former is inspired by the Q-Former in BLIP-2~\cite{li2023blip} and augments its content queries $q_{con}$ with sentence queries $q_{sen}$ and box queries $q_{box}$. $q_{sen}$ and $q_{box}$ are obtained by transforming the corresponding prompt features $G_{sen}$ and $G_{box}$ with two separate linear layers. We add $q_{sen}$ and $q_{box}$ to $q_{con}$ to incorporate semantic and positional cues~\cite{liu2022dab}. Note that the use of different types of queries not only enables our method to adapt to a variety of video tasks but also explicitly integrates guidance information from prompts into the visual features.

With this, we begin by splitting the video features $F$ along the temporal dimension, resulting in a sequence of frame features $\{ F_{i}\}_{i=1}^{T^{f}}$, and then send them to MQ-former in parallel. Within the MQ-former, the summed queries interact with one another, and $F_{i}$, through self-attention and cross-attention in an iterative manner, which integrates the frame features into the compact queries. Subsequently, we feed the per-frame queries to a transformer layer~\cite{dosovitskiy2021an} for temporal modeling, yielding $Q \in \mathcal{R}^{T^{f}N_{q} \times C_{q}}$, where $N_{q}$ is the number of queries and set to 32 following the configuration in BLIP-2~\cite{li2023blip}, $C_{q}$ represents the feature dimension.

\vspace{0.02in}
\noindent \textbf{Visual Translator.} Alignment between video and textual representations is extremely important to ensure that the output of our model is intrinsically relevant to the video content. To accomplish this, we input $Q$ to a Multi-Layer Perceptron (MLP) layer to project it to the textual embedding space, thereby aligning its dimension with the prompt features $G$. After this, they are concatenated along the sequence dimension to obtain the multimodal tokens $M \in \mathcal{R}^{(L^{g} + T^{f}N_{q}) \times C^{g}}$.

\vspace{0.02in}
\noindent \textbf{Video-grounded Token Decoding.} Finally, we employ a token decoder to predict a sequence of tokens based on $M$. The architecture of our token decoder is similar to popular language decoders~\cite{vas2017attention,lewis2019bart}, with causal self-attention for autoregressive token generation. 

\subsection{Unified Training and Inference}
\label{subsec:obj}
\textbf{Training}. Conditioned on $M$, \system is trained to maximize the log-likelihood between the predicted tokens $\hat{y}$ and the target tokens $y$ with cross-entropy loss:
\begin{equation}
    {\rm maximize} \sum_{k=1}^{\textit{L}} \texttt{log} \, \rm{P}(\hat{y}_{k} | M, y_{1:k-1}),
\end{equation}
where $\rm{P}$ denotes the softmax probability and $L$ is the length of $y$. Note that the output of various video tasks could be represented as a sequence of tokens in the unified vocabulary introduced in Sec.~\ref{subsec:tokenization}.

\vspace{0.05in}
\noindent \textbf{Inference}. During inference, we predict each token according to the model likelihood, \ie, ${P}(y_{k} | M, y_{1:k-1})$, and employ the beam search strategy~\cite{freitag2017beam} since it leads to the better performance than $\rm{argmax}$ sampling or nucleus sampling~\cite{holtzman2019curious}. Similar to language models, the end of sequence generation is indicated by an $\rm{EOS}$ token. The event segments for dense video captioning and bounding boxes for visual object tracking could be easily obtained by de-quantizing the \textit{time} or \textit{box tokens}.

%% file: sec/4_experiment.tex
\section{Experiments}
\label{sec:exp}

\begin{table}[t]
\small
\centering
\caption{Comparison with state-of-the-art video action recognition methods. Note that for MoViNet, we report the best results on both datasets, \ie, A6 on K400 and A3 on SSV2.}
\vspace{-0.1in}
\setlength{\tabcolsep}{2.5pt}
\renewcommand{\arraystretch}{1.1}
\begin{tabular*}{\linewidth}{lc|ccc|cc}
\toprule
\multirow{2}*{\textbf{Method}} && \multicolumn{2}{c}{\textbf{K400}} && \multicolumn{2}{c}{\textbf{SSV2}} \\
~ && \# Frames & Top1 && \# Frames & Top1 \\
\midrule
I3D~\cite{kay2017kinetics} && N/A & 72.1 && - & - \\
R(2+1)D-TS~\cite{tran2018closer} && N/A & 73.9 && - & - \\
SlowFast~\cite{feichtenhofer2019slowfast} && 8 $\times$ 3 $\times$ 10 & 77.9 && - & - \\
ip-CSN~\cite{tran2019video} && 32 $\times$ 3 $\times$ 10 & 79.2 && - & - \\
X3D-XL~\cite{feichtenhofer2020x3d} && 16 $\times$ 3 $\times$ 10 & 79.1 && - & - \\
SlowFast+NL~\cite{feichtenhofer2019slowfast} && 16 $\times$ 3 $\times$ 10 & 79.8 && - & - \\
CorrNet~\cite{wang2020video} && 32 $\times$ 3 $\times$ 10 & 81.0 && - & - \\
MoViNet~\cite{kondratyuk2021movinets} && 120 $\times$ 1 $\times$ 1 & 81.5 && 120 $\times$ 1 $\times$ 1 & 64.1 \\
\midrule
ViT-B-VTN~\cite{neimark2021video} && 250 $\times$ 1 $\times$ 1 & 78.6 && - & - \\
MViT-B~\cite{fan2021multiscale} && 32 $\times$ 1  $\times$ 5 & 80.2 && 64 $\times$ 3 $\times$ 1 & 67.7 \\
XViT~\cite{bulat2021space} && 16 $\times$ 3  $\times$ 1 & 80.2 && 32 $\times$ 3 $\times$ 1 & 65.4 \\
ViViT-L~\cite{arnab2021vivit} && 16 $\times$ 3  $\times$ 4 & 80.6 && 16 $\times$ 3 $\times$ 4 & 65.4 \\
TimeSformer-L~\cite{bertasius2021space} && 96 $\times$ 1 $\times$ 3 & 80.7 && 96 $\times$ 3 $\times$ 1 & 62.3\\
Mformer-HR~\cite{patrick2021keeping} && 16 $\times$ 3 $\times$ 10 & 81.1 && 16 $\times$ 3 $\times$ 1 & 67.1 \\
VideoSwin-B~\cite{liu2022video} && 32 $\times$ 3  $\times$ 4 & 82.7 && 32 $\times$ 3 $\times$ 1 & 69.6\\
UniFormer-B~\cite{li2022uniformer} && 32 $\times$ 1 $\times$ 4 & 82.9 && 32 $\times$ 3 $\times$ 1 & 71.2 \\
\rowcolor{Gray}
Ours && 32 $\times$ 3 $\times$ 4 & \textbf{83.6} && 32 $\times$ 3 $\times$ 1 & \textbf{71.3} \\
\bottomrule
\end{tabular*}
\vspace{-0.2in}
\label{tab:ar}
\end{table}

\subsection{Implementation Details}
\noindent \textbf{Datasets.} Our training corpus include action recognition datasets (Kinetics-400~\cite{kay2017kinetics} and Something-Something V2~\cite{goyal2017something}), clip captioning datasets (MSRVTT~\cite{xu2016msr} and MSVD~\cite{xu2017video}), video question answering datasets (MSRVTT~\cite{xu2016msr} and MSVD~\cite{xu2017video}), dense video captioning datasets (ActivityNet~\cite{caba2015activitynet}), and visual object tracking datasets (TrackingNet~\cite{muller2018trackingnet}, LaSOT~\cite{fan2019lasot}, GOT10K~\cite{huang2019got}). 

\vspace{0.05in}
\noindent \textbf{Model Instantiation.} We adopt VideoSwin pretrained on Kinetics-600~\cite{carreira2018short} as the video encoder, and initialize the language encoder and token decoder with pretrained Bart-base~\cite{lewis2019bart} model that owns $\sim$140M parameters. The number of time and box tokens are set to 300 and 1000, respectively. Following BLIP-2~\cite{li2023blip}, we adopt the same architecture of Bert-Base for our MQ-Former, which consists of 12 transformer layers with additionally inserted cross-attention blocks. The positional encodings are added to the outputs of MQ-Former to inject temporal information.

\vspace{0.05in}
\noindent \textbf{Training and Inference Procedures.} For the clip-based tasks, including action recognition (AR), clip captioning (CC), and video question answering (ViQA), we sample 32 frames randomly during training and uniformly during inference. For dense video captioning (DVP), we follow~\cite{yang2023vid2seq} to extract frames at 1FPS, and subsample or pad the frame sequence to 160 during both training and inference. For visual object tracking (VOT), we randomly sample two frames in a video sequence during training, following the common practice~\cite{chen2021transformer,wei2023autoregressive}.

We train our model for 50, 20, 50, and 500 epochs for AR, CC, ViQA, DVP, and VOT, respectively. Note that we follow~\cite{wei2023autoregressive,chen2023seqtrack} to train VOT for a longer time since the scale of tracking datasets is much larger. Different batch sizes are adopted, \ie, 64 for AR, 8 for CC, 256 for ViQA, 8 for DVP, and 16 for VOT. The model is optimized with the AdamW optimizer~\cite{loshchilov2017decoupled}, with an initial learning rate 5e-6 and decayed to 0 with the cosine scheduler. The frame resolution that we adopt is 224 $\times$ 224, augmented with random resized cropping and horizontal flipping. During inference, we average the logits of the generated tokens as the final score for AR to support multi-clip\&crop evaluation, and VOT for template update~\cite{wei2023autoregressive,chen2023seqtrack}. The threshold for VOT template update is 0.03.

\subsection{Main Results}
\noindent 1) Action Recognition, as one of the most representative video understanding tasks, aims to identify the action categories in a video. We evaluate the Top-1 accuracy of \system on commonly used datasets, including Kinetics-400 (K400)~\cite{kay2017kinetics} which consists of 306k short video clips of 400 action categories, and Something-Something V2 (SSV2)~\cite{goyal2017something} which comprises 220k videos of 174 categories. The comparison results with other methods are shown in Table~\ref{tab:ar}. \system achieves the best performance on both datasets, \ie, 83.6\% on K400 and 71.3\% on SSV2, surpassing VideoSwin~\cite{liu2022video} by 0.9 and 1.7, respectively. This highlights the advantage of our method.

\begin{table}[t]
\centering
\small
\caption{Comparison with state-of-the-art video captioning methods on MSRVTT and MSVD. Off-the-shelf object detectors are used for the results marked with $^{\dagger}$. }
\vspace{-0.1in}
\renewcommand{\arraystretch}{1.1}
\setlength{\tabcolsep}{1.8pt} % let TeX compute the intercolumn space
\begin{tabular*}{\linewidth}{l|ccccc|cccc}
\toprule
\multirow{2}*{\textbf{Method}} & \multicolumn{4}{c}{\textbf{MSRVTT}} && \multicolumn{4}{c}{\textbf{MSVD}} \\
~ & B@4 & M & R & C && B@4 & M & R & C  \\
\midrule
PickNet~\cite{chen2018less} & 41.3 & 27.7 & 59.8 & 44.1 && 52.3 & 33.3 & 69.6 & 76.5  \\
SibNet~\cite{liu2018sibnet} & 40.9 & 27.5 & 60.2 & 47.5 && 54.2 & 34.8 & 71.7 & 88.2 \\
OA-BTG$^{\dagger}$~\cite{zhang2019object} & 41.4 & 28.2 & - & 46.9 && 56.9 & 36.2 & - & 90.6 \\
GRU-EVE$^{\dagger}$~\cite{aafaq2019spatio} & 38.3 & 28.4 & 60.7 & 48.1 && 47.9 & 35.0 & 71.5 & 78.1 \\
MGSA~\cite{chen2019motion} & 42.4 & 27.6 & - & 47.5 && 53.4 & 35.0 & - & 86.7 \\
POS+CG~\cite{wang2019controllable} & 42.0 & 28.2 & 61.6 & 48.7 && 52.5 & 34.1 & 71.3 & 88.7 \\
POS+VCT~\cite{hou2019joint} & 42.3 & 29.7 & 62.8 & 49.1 && 52.8 & 36.1 & 71.8 & 87.8 \\
SAAT~\cite{zheng2020syntax} & 39.9 & 27.7 & 61.2 & 51.0 && 46.5 & 33.5 & 69.4 & 81.0 \\
STG-KD$^{\dagger}$~\cite{pan2020spatio} & 40.5 & 28.3 & 60.9 & 47.1 && 52.2 & 36.9 & 73.9 & 93.0 \\
PMI-CAP~\cite{chen2020learning} & 42.1 & 28.7 & - & 49.4 && 54.6 & 36.4 & - & 95.1 \\ 
ORG-TRL$^{\dagger}$~\cite{zhang2020object} & 43.6 & 28.8 & 62.1 & 50.9 && 54.3 & 36.4 & 73.9 & 95.2 \\
OpenBook~\cite{zhang2021open} & 42.8 & 29.3 & 61.7 & 52.9 && - & - & - & - \\
SwinBERT~\cite{lin2022swinbert} & 41.9 & 29.9 & 62.1 & 53.8 && 58.2 & 41.3 & 77.5 & 120.6 \\
\rowcolor{Gray}
Ours & \textbf{44.3} & \textbf{29.9} & \textbf{62.7} & \textbf{56.6} && \textbf{59.7} & \textbf{42.2} & \textbf{78.1} & \textbf{122.5} \\
\bottomrule
\end{tabular*}
\vspace{-0.1in}
\label{tab:vc}
\end{table}

\begin{table}[t]
\small
\centering
\caption{Accuracy (\%) of ViQA on MSRVTT and MSVD, Pre VLData: pertaining vision-language data.}
\vspace{-0.1in}
\setlength{\tabcolsep}{5.8pt}
\renewcommand{\arraystretch}{1.1}
\begin{tabular*}{\linewidth}{l | c | cc}
\toprule
\textbf{Method} & \textbf{PreTrain VLData} & \textbf{MSRVTT} & \textbf{MSVD} \\
\midrule
ClipBERT~\cite{lei2021less} & 5.6M & 37.4 & - \\
CoMVT~\cite{seo2021look} & 100M & 39.5 & 42.6 \\
JustAsk~\cite{yang2021just} & 69M & 41.5 & 46.3 \\
ALIPRO~\cite{li2022align} & 5.5M & 42.1 & 45.9 \\
OmniVL~\cite{wang2022omnivl} & 18M & 44.1 & 51.0 \\
\midrule
HCRN~\cite{le2020hierarchical} & - & 35.6 & 36.1 \\
JustAsk~\cite{yang2021just} & - & 39.6 & 41.2 \\
\rowcolor{Gray}
Ours & - & \textbf{42.3} & \textbf{47.7} \\
\bottomrule
\end{tabular*}
\vspace{-0.2in}
\label{tab:vidqa}
\end{table}

\begin{table}[t]
\small
\centering
\caption{Dense captioning on the ActivityNet Captions validation set. $^{*}$ denotes pretraining on large-scale video-language dataset YT-Temporal-1B~\cite{zellers2022merlot}.}
\vspace{-0.1in}
\renewcommand\arraystretch{1.1}
\setlength{\tabcolsep}{2.8pt}
\begin{tabular*}{\linewidth}{lc | cccc | ccc | c}
\toprule
\multirow{2}*{\textbf{Method}} && \multicolumn{3}{c}{\textbf{Captioning}} && \multicolumn{2}{c}{\textbf{Event Loc.}} && \textbf{Overall} \\
~ && B4 & M & C && R & P && SODA$\_$c \\
\midrule
DCE~\cite{krishna2017dense} && 0.17 & 5.69 & 12.43 && - & - && - \\
DVC~\cite{li2018jointly} && 0.73 & 6.93 & 12.61 && - & - && - \\
TDA-CG~\cite{wang2018bidirectional} && 1.31 & 5.86 & 7.99 && - & - && - \\
SDVC~\cite{mun2019streamlined} && - & 6.92 & - && 55.58 & 57.57 && - \\
PDVC~\cite{wang2021end} && 1.65 & 7.50 & 25.87 && 55.42 & 58.07 && 5.3 \\
UEDVC~\cite{zhang2022unifying} && - & - & - && \textbf{59.00} & 60.32 && 5.5 \\
\midrule
Vid2seq~\cite{yang2023vid2seq} && - & - & 18.80 && - & - && 5.4 \\
\textcolor{gray}{Vid2seq$^{*}$~\cite{yang2023vid2seq}} && \textcolor{gray}{-} & \textcolor{gray}{8.50} & \textcolor{gray}{30.10} && \textcolor{gray}{52.70} & \textcolor{gray}{53.90} && \textcolor{gray}{5.8} \\
\rowcolor{Gray}
Ours && \textbf{1.73} & \textbf{7.54} & \textbf{26.00} && 45.08 & \textbf{60.43} && \textbf{5.6} \\
\bottomrule
\end{tabular*}
\vspace{-0.2in}
\label{tab:dvp}
\end{table}

\vspace{0.05in}
\noindent 2) Video Captioning expects the model to generate a textual description for a given video, which simultaneously evaluates the visual comprehension and text generation capability of our method. MSRVTT~\cite{xu2016msr} and MSVD~\cite{chen2011collecting}, two large-scale open domain video captioning datasets, are adopted and the results are shown in Table~\ref{tab:vc}. We can see that \system outperforms existing models by a clear margin (+2.8 and +1.9 in terms of CIDEr on MSRVTT and MSVD), even if several of them, \eg, OA-BTG~\cite{zhang2019object} and ORG-TRL~\cite{zhang2020object}, leverage object detector~\cite{he2017mask,girshick2015fast} to extract object information in an offline manner.

\vspace{0.02in}
\noindent 3) Video Question Answering aims to answer a natural language question based on the video content. We compare the accuracy of \system with other ViQA models on MSRVTT~\cite{xu2016msr} and MSVD~\cite{xu2017video} in Table~\ref{tab:vidqa}. The results demonstrate that \system outperforms both QA-specific methods, \eg, JustAsk~\cite{yang2021just}, and pertaining methods, \eg, ALIPRO~\cite{li2022align}, showcasing the effectiveness of our method for complex multimodal reasoning.

\vspace{0.02in}
\noindent 4) Dense Video Captioning localizes the events in an untrimmed video and generates the corresponding text descriptions for them. Following the practice of previous methods~\cite{wang2021end,yang2023vid2seq}, we evaluate \system in three aspects: 1) the average precision (P), average recall (R) across IOU at {0.3, 0.5, 0.7, 0.9} and their harmonic mean for localization. 2) BLEU4 (B4), METEOR (M), and CIDEr (C) for dense captioning. 3) SODA$\_$c for an overall evaluation. The results are reported in Table~\ref{tab:dvp}.

Traditional methods, including both two-stage (\eg, DVC~\cite{li2018jointly}, SDVC~\cite{mun2019streamlined}), and one-stage models (\eg, PDVC~\cite{wang2021end}, UEDVC~\cite{zhang2022unifying}), all employ the pre-extracted features from video backbones~\cite{kay2017kinetics} without end-to-end training. Compared to them, \system achieves better results on all the metrics, except for Recall. Our underperformance on recall is because traditional methods always apply a fixed number of localization heads to get a large number of false-positive predictions, \eg, 100 for SDVC~\cite{mun2019streamlined}. Vid2Seq~\cite{yang2023vid2seq} is the first end-to-end framework for dense video captioning. We can see that our method, although slightly inferior to their pre-trained model on YT-Temporal-1B, can significantly outperform them without large-scale pretraining, \ie, 18.80 \vs 26.00 in terms of CIDEr. A detailed comparison between \system and Vid2seq can be found in the appendix.

\begin{table}[t]
\small
\centering
\caption{Comparisons with the visual object tracking models on LaSOT and TrackingNet. }
\vspace{-0.1in}
\renewcommand\arraystretch{1.1}
\setlength{\tabcolsep}{3.5pt} % let TeX compute the intercolumn space
\begin{tabular*}{\linewidth}{lc | cccc | cccc }
\toprule
\multirow{2}*{\textbf{Method}} && \multicolumn{3}{c}{\textbf{LaSOT}} && \multicolumn{3}{c}{\textbf{TrackingNet}}  \\
~ && Suc & P$_{\rm{norm}}$ & P && Suc & P$_{\rm{norm}}$ & P \\
\midrule
SiamFC~\cite{bertinetto2016fully} && 33.6 & 42.0 & 33.9 && 57.1 & 66.3 & 53.3 \\
ATOM~\cite{danelljan2019atom} && 51.5 & 57.6 & 50.5 && 70.3 & 77.1 & 64.8 \\
SiamPRN++~\cite{li2019siamrpn++} && 49.6 & 56.9 & 49.1 && 73.3 & 80.0 & 69.4 \\
DiMP~\cite{bhat2019learning} && 56.9 & 65.0 & 56.7 && 74.0 & 80.1 & 68.7 \\
KYS~\cite{bhat2020know} && 55.4 & 63.3 & - && 74.0 & 80.0 & 68.8 \\
Ocean~\cite{zhang2020ocean} && 56.0 & 65.1 & 56.6 && - & - & - \\
AutoMatch~\cite{zhang2021learn} && 58.2 & - & 59.9 && 76.0 & - & 72.6 \\
PrDiMP~\cite{danelljan2020probabilistic} && 59.8 & 68.8 & 60.8 && 75.8 & 81.6 & 70.4 \\
TrDiMP~\cite{wang2021transformer} && 63.9 & - & 61.4 && 78.4 & 83.3 & 73.1 \\
Siam R-CNN~\cite{voigtlaender2020siam} && 64.8 & 72.2 & - && 81.2 & 85.4 & 80.0 \\
TransT~\cite{chen2021transformer} && 64.9 & 73.8 & 69.0 && 81.4 & 86.7 & 80.3 \\
Unicorn~\cite{yan2022towards} && 68.5 & 76.6 & 74.1 && 83.0 & 86.4 & 82.2 \\
KeepTrack~\cite{mayer2021learning} && 67.1 & 77.2 & 70.2 && - & - & -\\
STARK~\cite{yan2021learning} && 67.1 & 77.0 & - && 82.0 & 86.9 & - \\
AiATrack~\cite{gao2022aiatrack} && - & 79.4 & 73.8 && - & 87.8 & 80.4 \\
OSTrack~\cite{ye2022joint} && - & 78.7 & 75.2 && - & 87.8 & 82.0 \\
MixFormer~\cite{cui2022mixformer} && 69.2 & 78.7 & 74.7 && 83.1 & 88.1 & 81.6\\
SeqTrack~\cite{chen2023seqtrack} && 69.9 & 79.7 & 76.3 && 83.3 & 88.3 & 82.2  \\
ARTrack~\cite{wei2023autoregressive} && 70.4 & 79.5 & 76.6 && 84.2 & 88.7 & 83.5 \\
UNINEXT~\cite{yan2023universal} && 72.4 & \textbf{80.7} & \textbf{78.9} && \textbf{85.1} & 88.2 & \textbf{84.7} \\
\rowcolor{Gray}
Ours && \textbf{70.8} & 79.6 & \textbf{76.9} && 83.8 & \textbf{88.9} & 83.2 \\
\bottomrule
\end{tabular*}
\label{tab:sot}
\vspace{-0.2in}
\end{table}

\vspace{0.02in}
\noindent 5) Visual Object Tracking estimates the trajectory of a target object given its position in the first frame, which requires a fine-grained understanding of spatial-temporal information. In Table~\ref{tab:sot}, we compare \system with other tracking models on two most representative datasets, LaSOT~\cite{fan2019lasot} and TrackingNet~\cite{muller2018trackingnet}. Success (Suc), precision (P), and normalized precision (P$_{\rm{norm}}$) are reported. It is worth mentioning that although SeqTrack~\cite{chen2023seqtrack} and ARTrack~\cite{wei2023autoregressive} also employ the autoregressive framework for object tracking, \system differs from them in twofold aspects. Firstly, we perform tracking on the complete frame, instead of a cropped region. Second, we encode the reference box to the visual feature of the tracking frame through box queries, rather than just using it as a prompt for the token decoder. It can be observed that \system achieves excellent performance on both LaSOT and TrackingNet, \ie, 79.6 and 88.9 in terms of P$_{\rm{norm}}$, which beats most of the previous SOTA methods.

\subsection{Ablation Studies}
\noindent \textbf{Analysis of Different Components in \system.} In Table ~\ref{tab:components}, we conduct ablation experiments to study the effects of the core components in \system: 1) text $\&$ box queries in Mixed Qformer: different queries are the core design of our method to adapt to different video tasks and inject reference information into the frame feature. It can be seen from the 1st and 2nd rows that they improve the VQA and VOT performance by 1.9 and 1.4, respectively. 2) temporal encoder: comparing the results in the 3rd and 5th rows, it is evident that the temporal encoder brings remarkable performance gains on all the tasks, validating the temporal modeling is important for video understanding. 3) initializing token decoder with Bart~\cite{lewis2019bart}: the results in row 4 demonstrate that the initialization of the token decoder has a greater impact on captioning tasks, stemming from the fact that the training objectives of captioning tasks are inherently more aligned with the pretraining of the token decoder.

\begin{table}[t]
\small
\centering
% \vspace{-0.05in}
\caption{Ablation studies on different components of \system.}
\vspace{-0.1in}
\renewcommand{\arraystretch}{1.1}
\begin{tabular*}{\linewidth}{@{\makebox[1em][l]{\rownumber\space}} | lc|ccccc }
\toprule
\textbf{Model} && \textbf{AR} & \textbf{CC} & \textbf{ViQA} & \textbf{DVP} & \textbf{VOT} 
\gdef\rownumber{\stepcounter{magicrownumbers}\arabic{magicrownumbers}} \\
\midrule
w/o TextQuery && 83.4 & 56.5 & 40.4 & 5.6 & 79.2 \\
w/o BoxQuery && - & - & - & - & 78.2 \\
w/o TemEnc && 82.5 & 53.3 & 41.7 & 5.1 & 77.6 \\
w/o LangInit && 81.7 & 44.4 & 39.7 & 4.5 & 79.0 \\
\rowcolor{Gray}
Ours && \textbf{83.6} & \textbf{56.6} & \textbf{42.3} & \textbf{5.6} & \textbf{79.6} \\
\bottomrule
\end{tabular*}
\vspace{-0.1in}
\label{tab:components}
\end{table}

\begin{table}[t]
\small
\centering
\renewcommand\arraystretch{1.0}
\caption{Open-vocabulary results on HMDB-51 and UCF101.}
\vspace{-0.1in}
\begin{tabular*}{\linewidth}{@{\extracolsep{\fill}}lc | cc | cc @{}}
\toprule
\textbf{Method} && \textbf{Train} && \textbf{HMDB-51} & \textbf{UCF101} \\
\midrule
ASR~\cite{wang2017alternative} && K400 && 21.8$\pm$ 0.9 & 24.4$\pm$1.0 \\
ZSECOC~\cite{qin2017zero} && K400 && 22.6$\pm$1.2 & 15.1$\pm$1.7 \\
UR~\cite{zhu2018towards} && K400 && 24.4$\pm$1.6 & 17.5$\pm$1.6 \\
E2E~\cite{brattoli2020rethinking} && K400 && 32.7 & 48.0 \\
Ours && K400 && 26.3 & 32.0 \\
\bottomrule
\end{tabular*}
% \vspace{-0.1in}
\label{tab:open}
\end{table}

\vspace{0.05in}
\noindent \textbf{Open-vocabulary Action Recognition:} Compared to the traditional classifier-based methods, \system is more flexible in adapting to the open-vocabulary (OV) setting by appending the category names to the input textual prompt. As shown in Table~\ref{tab:open}, \system achieves competitive results than existing OV methods without cumbersome designs.

\begin{figure}[t]
\centering
\includegraphics[width=\linewidth]{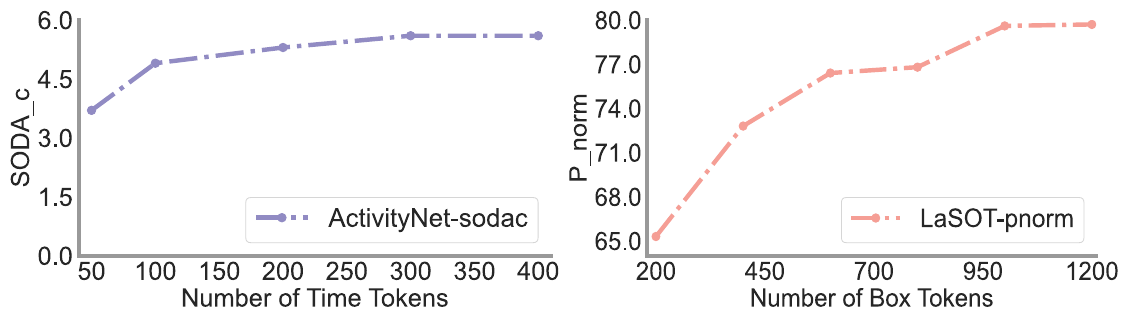}
\vspace{-0.3in}
\figcaption{Comparison between joint and separate training.}
\label{fig:tokens}
\vspace{-0.1in}
\end{figure}

\vspace{0.05in}
\noindent \textbf{Number of Time and Box Tokens.} We further try different numbers of time ($N_{t}$) and box ($N_{b}$) tokens on the localization tasks. As shown in Figure~\ref{fig:tokens}, for both types of tokens, increasing the number could first improve the results since the quantization error is reduced accordingly, and finally converges when $N_{t} \geq 300$ and  $N_{b} \geq 1000$.

\begin{figure*}[t]
\centering
\includegraphics[width=\linewidth]{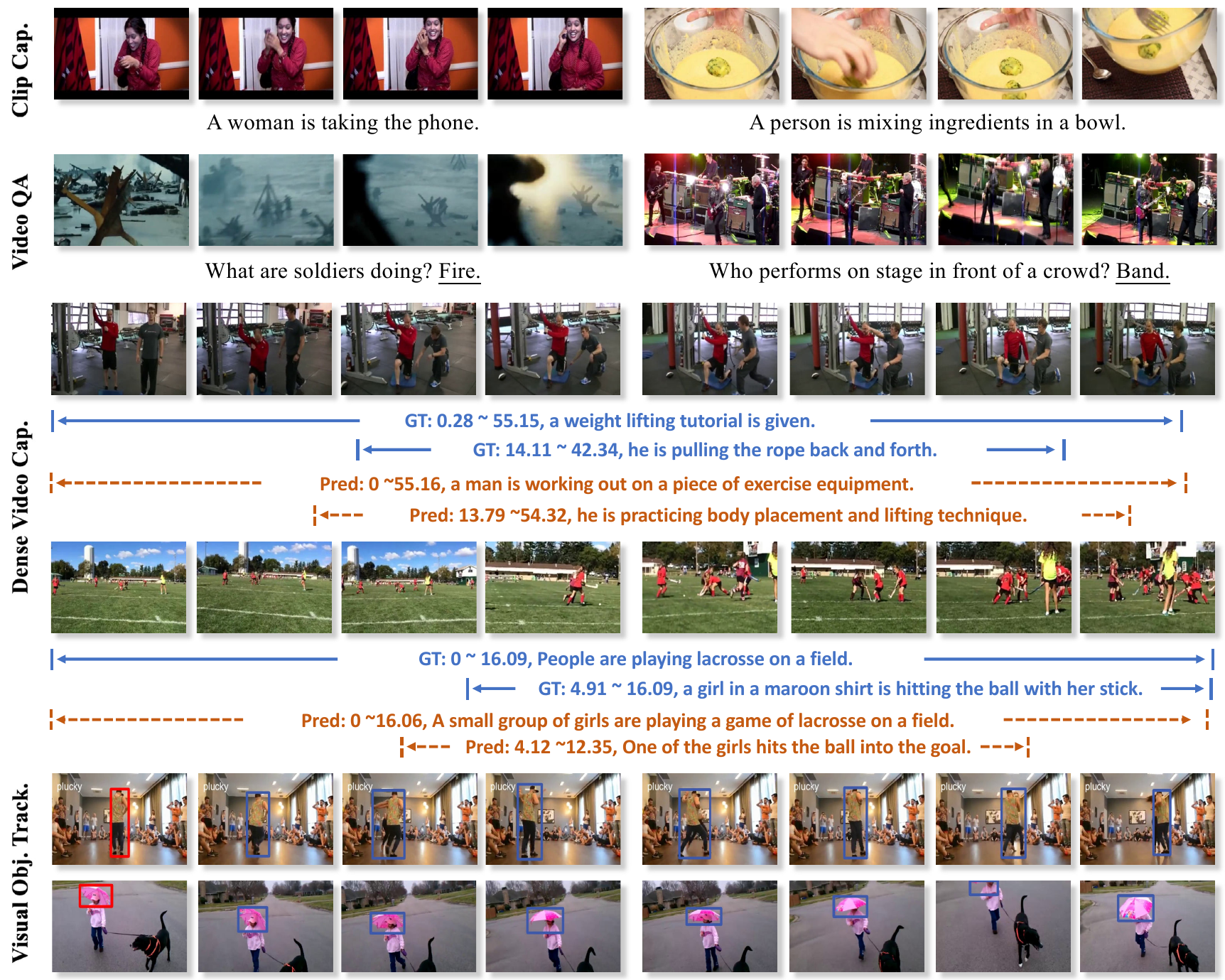}
\vspace{-0.2in}
\caption{Visualization of the predictions by \system on different video understanding tasks. From top to down, we show the clip captioning, video question answering, dense video captioning, and visual object tracking visualization results, respectively.}
\label{fig:visualization}
\vspace{-0.15in}
\end{figure*}

\subsection{Visualizations}
We visualize the predictions of \system on various video understanding tasks in Figure~\ref{fig:visualization}. From the top two rows, we can see that \system could not only generate accurate and natural captions for videos but also answer questions regarding the characters or activities in the video, showcasing its cross-modal modeling capability. In addition, \system also excels in spatial-temporal localization. The results in 3rd and 4th rows show that it could detect different types of events in videos precisely and produce vivid descriptions for them. Moreover, \system also exhibits remarkable robustness against occlusions and variations in object tracking. These visualizations underscore the versatility and effectiveness of \system across a wide range of video tasks.

%% file: sec/5_conclusion.tex
\section{Conclusion}
\label{sec:conclusion}
This paper introduced \system, a generative framework for universal video understanding. We defined a unified output space for different video tasks by supplementing the vocabulary of language models with special \textit{time} and \textit{box tokens}. With this, a wide spectrum of video tasks, including action recognition, clip captioning, video question answering, dense video captioning, and visual object tracking, could be formulated as a video-grounded token generation process, and further, addressed within an encoder-decoder architecture. Extensive experiments on seven prominent video benchmarks showcased the superior video understanding capability and versatility of \system.

Despite the promising results achieved, the joint training performance of \system exhibited some degradation in the spatial-temporal localization tasks compared to separate training. In the future, we will explore more advanced training and optimization strategies on multiple datasets and tasks, to further improve the overall performance and robustness of our method.

\noindent \textbf{Acknowledgement} This project was supported by NSFC under Grant No. 62032006 and No. 62102092.